\documentclass[10pt, a4paper]{article}

\usepackage[final]{lrec2026} 
\usepackage{multirow}
\usepackage{caption}
\usepackage{makecell}
\usepackage{graphicx}%
\usepackage{multirow}%
\usepackage{amsmath,amssymb,amsfonts}%
\usepackage{amsthm}%
\usepackage{mathrsfs}%
\usepackage[title]{appendix}%
\usepackage{xcolor}%
\usepackage{textcomp}%
\usepackage{manyfoot}%
\usepackage{booktabs}%
\usepackage{algorithm}%
\usepackage{algorithmicx}%
\usepackage{algpseudocode}%
\usepackage{listings}%
\usepackage{microtype}%
\usepackage{array}
\usepackage{lipsum}
\usepackage{longtable}
\usepackage{float}
\usepackage{adjustbox}
\usepackage[T5]{fontenc} 
\usepackage{mathptmx}
\usepackage{subfigure}
\usepackage{amsmath}
\usepackage{pifont}
\usepackage{tabularx}
\usepackage{arydshln}

\title{ViWikiFC: Fact-Checking for Vietnamese Wikipedia-Based Textual Knowledge Source}

\name{Hung Tuan Le, Long Truong To, Manh Trong Nguyen, Kiet Van Nguyen$^{*}$} 

\address{University of Information Technology, Ho Chi Minh City, Vietnam \\
    Vietnam National University, Ho Chi Minh City, Vietnam\\
        \{21520250, 21521101, 21520343\}@gm.uit.edu.vn, kietnv@uit.edu.vn\\}

\abstract{
Fact-checking is essential due to the explosion of misinformation in the media ecosystem. Although false information exists in every language and country, most research to solve the problem has mainly concentrated on huge communities like English and Chinese. Low-resource languages like Vietnamese are necessary to explore corpora and models for fact verification. To bridge this gap, we construct ViWikiFC, the first manually annotated open-domain corpus for \textbf{Vi}etnamese \textbf{Wiki}pedia \textbf{F}act \textbf{C}hecking more than 20K claims generated by converting evidence sentences extracted from Wikipedia articles. We analyze our corpus through many linguistic aspects, from the new dependency rate, the new n-gram rate, and the new word rate. We conducted various experiments for Vietnamese fact-checking, including evidence retrieval and verdict prediction. BM25 and InfoXLM$_{Large}$ achieved the best results in two tasks, with BM25 achieving an accuracy of 88.30\% for SUPPORTS, 86.93\% for REFUTES, and only 56.67\% for the NEI label in the evidence retrieval task. InfoXLM$_{Large}$ achieved an F$_1$ score of 86.51\%. Furthermore, we also conducted a pipeline approach, which only achieved a strict accuracy of 67.00\% when using InfoXLM$_{Large}$ and BM25. These results demonstrate that our dataset is challenging for the Vietnamese language model in fact-checking tasks.
 \\ \newline  }

\begin{document}

\maketitleabstract
\section{Introduction}
\label{sec:Introduction}

\renewcommand{\thefootnote}{\fnsymbol{footnote}}
\footnotetext[1]{Corresponding author.}
\renewcommand{\thefootnote}{\arabic{footnote}}

\begin{table*}[!ht]
\centering
\resizebox{1\textwidth}{!}{
\begin{tabular}{lcccccc}
\hline
\multicolumn{1}{c}{\textbf{Corpus}} & \textbf{Claims} & \textbf{Language} & \textbf{Source} & \textbf{Domain} & \textbf{Annotated Evidence} & \textbf{Annotated Label} \\ \hline
PolitiFact (2014) & 106 & English & PolitiFact & Politics & \textcolor{red}{\ding{55}} & \textcolor{red}{\ding{55}} \\
Liar (2017) & 12,836 & English & PolitiFact & Politics & \textcolor{red}{\ding{55}} & \textcolor{red}{\ding{55}} \\
FEVER (2018) & 185,445 & English & Wiki & Multiple & \textcolor{blue}{\ding{51}} & \textcolor{blue}{\ding{51}} \\
PUBHEALTH (2020) & 11,832 & English & Fact-Checking/News & Health & \textcolor{red}{\ding{55}} & \textcolor{red}{\ding{55}} \\
HOVER (2020) & 26,171 & English & Wiki & Multiple & \textcolor{blue}{\ding{51}} & \textcolor{blue}{\ding{51}} \\
TabFact (2020) & 92,283 & English & Wiki & Multiple & \textcolor{red}{\ding{55}} & \textcolor{blue}{\ding{51}} \\
InfoTabs (2020) & 23,738 & English & Wiki & Multiple & \textcolor{red}{\ding{55}} & \textcolor{blue}{\ding{51}} \\
ANT (2020) & 4,547 & Arabic & News & Multiple & \textcolor{red}{\ding{55}} & \textcolor{red}{\ding{55}} \\
FakeCovid (2020) & 5,182 & Multilingual (3) & Fact Check & Health & \textcolor{red}{\ding{55}} & \textcolor{red}{\ding{55}} \\
X-Fact (2021) & 31,189 & Multilingual (25) & Fact Check & Multiple & \textcolor{blue}{\ding{51}} & \textcolor{red}{\ding{55}} \\
VitaminC (2021) & 488,904 & English & Wiki & Multiple & \textcolor{red}{\ding{55}} & \textcolor{blue}{\ding{51}} \\
DanFEVER (2021) & 6,407 & Danish & Wiki & Multiple & \textcolor{blue}{\ding{51}} & \textcolor{blue}{\ding{51}} \\
CHEF (2022) & 10,000 & Chinese & Internet & Multiple & \textcolor{blue}{\ding{51}} & \textcolor{blue}{\ding{51}} \\ \hline
\textbf{ViWikiFC} & \textbf{20,916} & \textbf{Vietnamese} & \textbf{Wiki} & \textbf{Multiple} & \textcolor{red}{\ding{55}} & \textcolor{blue}{\ding{51}} \\ \hline
\end{tabular}
}
\caption{Overall statistics of fact-checking corpora.}
\label{tab:Corpus Information}
\end{table*}

Following the increase in the amount of information, the lack of strict policies when spreading information has led to misinformation and disinformation on social media. This can cause conflict and manipulate the group of people. To minimize these impacts, organizations like PolitiFact and FactCheck.org play an essential role in online Fact Verification by manually verifying claims based on different sources of evidence. However, manual verification is time-consuming and insufficient given the speed of information updates on social media.

Automatic Fact-Checking is a complicated task that can be separated into four sub-tasks: claim detection, evidence retrieval, verdict prediction, and justification production \cite{guo-etal-2022-survey}. While there are many works on improving fact-checking systems \cite{lewis2020retrieval,maillard-etal-2023-small,pan-etal-2023-fact} and large-scale corpora for various languages such as NELA \cite{horne2018sampling}, FakeCovid \cite{shahi2020fakecovid}, FEVER \cite{thorne-etal-2018-fever}, and VitaminC \cite{schuster-etal-2021-get}, Vietnamese remains a low-resource language for NLP, especially in automatic fact-checking.

To contribute to Vietnamese NLP research, we present ViWikiFC: the first large-scale, open-domain corpus for Vietnamese Fact-Checking on Wikipedia. The corpus consists of 20,916 manually annotated claims based on evidence from Wikipedia pages. Following FEVER \cite{thorne-etal-2018-fever}, our corpus has three label classes: SUPPORTS, REFUTES, and NOTENOUGHINFORMATION (NEI). The evidence is manually rewritten into three types of claims, making them more realistic and semantically diverse than FEVER's approach. We also follow ViNLI \cite{huynh-etal-2022-vinli} to create double claims with different meanings for each label from one evidence.

Annotators who are native Vietnamese speakers with education beyond high school were carefully trained to ensure consistency. We conducted experiments on evidence retrieval using TF-IDF, BM25, and Vietnamese-SBERT \cite{10.1007/978-3-031-15063-0_40}, and on verdict prediction using CBOW \cite{mikolov2013distributed}, BiLSTM \cite{graves2005framewise}, multilingual models (mBERT \cite{kenton2019bert}, XLM-R \cite{conneau-etal-2020-unsupervised}, InfoXLM \cite{chi-etal-2021-infoxlm}), and Vietnamese-specific models (PhoBERT \cite{nguyen-tuan-nguyen-2020-phobert}, ViDeBERTa \cite{tran-etal-2023-videberta}).

Contributions in this paper are described as follows.

\begin{itemize} 
\item Firstly, we propose the first fact-checking corpus on Vietnamese Wikipedia, comprising 20,916 claim sentences manually annotated based on 73 Wikipedia articles built with a consensus agreement among annotators, achieving a 95.87\% Fleiss'k-agreement. 
\item Second, we conduct two experiments on two tasks, evidence retrieval and verdict prediction, while also developing a pipeline approach for fact-checking models, encompassing both neural network-based and pre-trained transformer-based models. 
\item Next, we analyze the corpus from various linguistic perspectives in the Vietnamese language to gain more insight into Vietnamese fact-checking, including the new n-gram rate, the new dependency rate, and the new word rate. 
\item Finally, our corpus and annotation tool are available for research purposes. 
\end{itemize}



\section{Related Work}
\label{sec:related work}
In this section, we review prior studies and resources related to fact-checking, focusing on corpora and baseline models that have shaped this research direction. Section \ref{subsec:Fact-Checking Corpus} introduces key datasets, while Section \ref{subsec:Fact-Checking baselines} presents the main methodological approaches.

\subsection{Fact-Checking Corpora}
\label{subsec:Fact-Checking Corpus}
We reviewed existing fact-checking datasets, summarized in Table \ref{tab:Corpus Information}, categorized by dataset size, language, source, domain, and annotation types.

The PolitiFact corpus \cite{vlachos-riedel-2014-fact} laid the foundation for modern fact-checking by defining three core tasks: claim detection, evidence retrieval, and verdict prediction. However, its limited size restricted model training. The LIAR dataset \cite{wang-2017-liar} expanded on this with over 12.8K samples from PolitiFact\footnote{https://www.politifact.com/} but lacked evidence annotations. FEVER \cite{thorne-etal-2018-fever} later introduced a large-scale corpus of 180K manually written claims based on Wikipedia\footnote{https://www.wikipedia.org/}, establishing an automated pipeline structure for verification. Subsequent corpora such as Snopes \cite{hanselowski-etal-2019-richly} and MultiFC \cite{augenstein-etal-2019-multifc} diversified sources and label types, while later datasets like HoVer \cite{jiang-etal-2020-hover} and VitaminC \cite{schuster-etal-2021-get} expanded evidence coverage and reasoning complexity.

For semi-structured data, TabFact \cite{2019TabFactA}, InfoTabs \cite{gupta-etal-2020-infotabs}, and FEVEROUS \cite{aly2021feverous} introduced fact verification using tables or hybrid text-table formats. Multilingual efforts have also emerged, including X-Fact \cite{gupta-srikumar-2021-x}, ANT \cite{khouja-2020-stance}, DanFEVER \cite{norregaard-derczynski-2021-danfever}, and CHEF \cite{hu-etal-2022-chef}. 

Despite this progress, Vietnamese NLP still lacks an open-domain, large-scale, and well-annotated fact-checking corpus. To address this gap, we constructed a Vietnamese fact-checking dataset and developed an accompanying verification pipeline using pre-trained Vietnamese language models.
\begin{figure*}[t]
    \centering
    \includegraphics[width=0.8\textwidth]{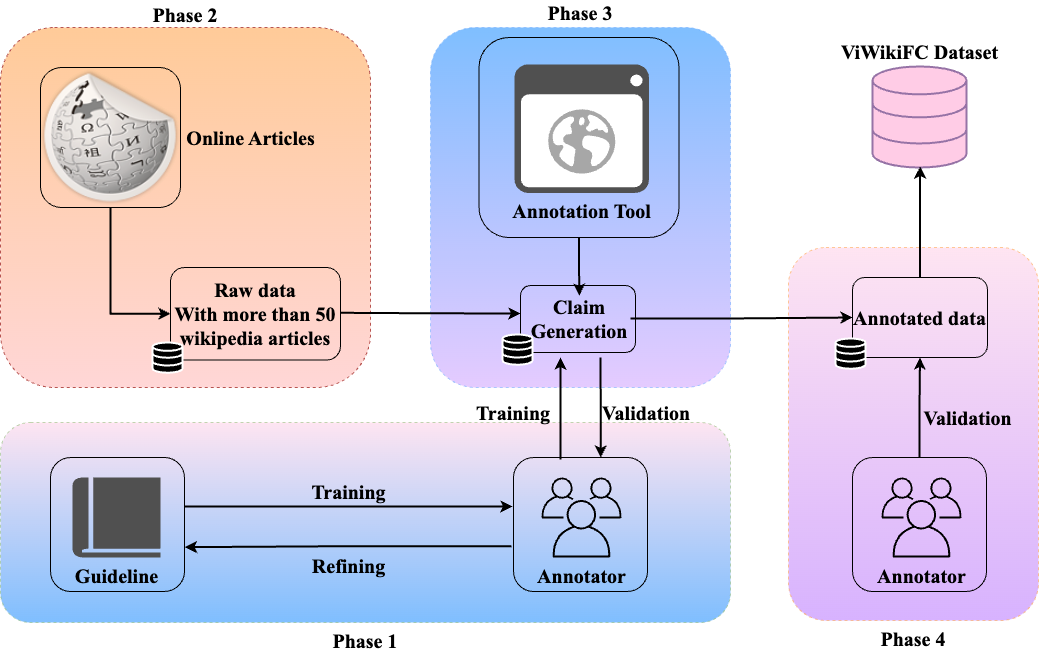}
    \caption{The creation process of the fact-checking corpus.}
    \label{fig:corpus creation}
\end{figure*}

\subsection{Fact-Checking Methods}
\label{subsec:Fact-Checking baselines}
Fact-checking typically involves four sub-tasks: claim detection, evidence retrieval, verdict prediction, and justification generation. Among these, evidence retrieval and verdict prediction are central, forming the core of our research.

Evidence retrieval involves finding relevant information from various sources to support or refute a claim. Traditional non-neural methods such as TF-IDF \cite{thorne-etal-2018-fever,hu-etal-2022-chef} and BM25 \cite{chen-etal-2017-reading} rely on lexical matching but are limited in semantic understanding. Neural approaches, such as Dense Passage Retrieval (DPR) \cite{karpukhin-etal-2020-dense}, use dense vector representations via dual-encoder architectures, achieving significant improvements in open-domain QA \cite{karpukhin-etal-2020-dense,chang2019pre,lee-etal-2019-latent} and fact-checking \cite{samarinas-etal-2021-improving,samarinas2020latent}. Cross-encoder re-rankers based on BERT and its variants \cite{nogueira2019passage,nogueira2019multi,wang2019multi} further enhance retrieval precision.

Verdict prediction determines the truthfulness of a claim based on retrieved evidence. It is typically formulated as a sentence-pair classification problem. Early approaches used CBOW \cite{mikolov2013distributed}, RNN \cite{ELMAN1990179}, or BiLSTM \cite{graves2005framewise}, while transformer-based models such as BERT \cite{devlin-etal-2019-bert} and its successors—RoBERTa \cite{DBLP:journals/corr/abs-1907-11692}, ALBERT \cite{lan2019albert}, and XLM-R \cite{conneau-etal-2020-unsupervised}—greatly improved performance. In Vietnamese, PhoBERT \cite{nguyen-tuan-nguyen-2020-phobert} was the first monolingual pre-trained model, trained on a 20GB corpus of Wikipedia and news data, followed by ViT5 \cite{phan-etal-2022-vit5} and ViDeBERTa \cite{tran-etal-2023-videberta}, which further enhanced representation quality for downstream tasks, including fact-checking.

\section{Corpus Creation}
\label{sec:Corpus Creation}
The corpus was created throughout four phases (see Figure \ref{fig:corpus creation}), including annotator recruitment and training (see Section \ref{subsec:Annotator Recruitment and Training}), evidence selection (see Section \ref{subsec:Evidence Selection}), claim generation (see Section \ref{subsec:Claim Generation}), and corpus validation (see Section \ref{subsec:Corpus Validation}). Furthermore, we also analyzed the corpus (see Section \ref{subsec:Corpus Analysis}) in terms of various linguistic aspects to gain a deeper understanding of the language characteristics of the corpus.
\subsection{Annotator Recruitment and Training}
\label{subsec:Annotator Recruitment and Training}
We recruited thirty Vietnamese native speakers with at least high school to college literacy to create claims based on given evidence sentences. All annotators were familiar with online reading and information searching. Before participating in the official annotation, they underwent a two-phase training process to ensure annotation quality, with agreement measured by the Fleiss $\kappa$ score \cite{Fleiss1971MeasuringNS}.

In the first phase, annotators were trained using detailed guidelines to understand claim-generation rules. They then annotated a small training corpus containing 50 evidence sentences labeled as \textit{SUPPORTS}, \textit{REFUTES}, or \textit{NOT ENOUGH INFORMATION}. Each claim-evidence pair was reviewed by one of three authors, and inter-author agreement above 0.90 was required for the annotator to proceed to the second phase. Annotators scoring below 0.90 were asked to review their mistakes and repeat the first phase using a new training set.

In the second phase, the thirty annotators were randomly divided into six groups of five. Each annotator labeled at least one dataset folder of 150 claim-evidence pairs, with labels concealed. Every group then re-labeled five folders produced by other groups, and inter-annotator agreement was again evaluated using Fleiss $\kappa$. Annotators achieving a consensus score of 0.95 or higher qualified for the official annotation process; otherwise, they repeated the first training phase. This procedure continued until all annotators achieved the required agreement threshold. During pilot annotations, annotators were encouraged to document ambiguous cases (e.g., polysemy, unclear entities) and note common labeling errors to refine the annotation guidelines.

\subsection{Evidence Selection}
\label{subsec:Evidence Selection}
We use Wikipedia as the primary data source to build the ViWikiFC corpus. We do not prioritize establishing the credibility of the data source, as, in reality, all data sources can sometimes present inaccuracies or omit information, either unintentionally or deliberately. Therefore, no source can be deemed completely trustworthy. The team considers the coverage of information in various aspects of life as a crucial factor. Wikipedia, being an open encyclopedia with more than 55 million articles on a wide array of topics, helps to build a corpus with greater comprehensiveness compared to constructing it solely from other sources, such as political information from Politifact\footnote{https://www.politifact.com/} or news articles from VnExpress\footnote{https://vnexpress.net/}. We extracted 3,812 evidence sentences from more than 1,479 paragraphs in 73 articles on Wikipedia, covering various topics such as history, geography, philosophy, and science.
\subsection{Claim Generation}
\label{subsec:Claim Generation}
Following the FEVER dataset \cite{thorne-etal-2018-fever}, we design three labels—SUPPORTS, REFUTES, and NOT ENOUGH INFORMATION (NEI)—to capture the different relationships between a claim and its supporting evidence. In practice, many claims cannot be conclusively verified or falsified due to incomplete or ambiguous information; therefore, we include the NEI label to better reflect these real-world scenarios. Annotators are instructed to compose claim sentences for each of the three labels as follows:

\begin{itemize}
    \item \textbf{SUPPORTS}: The claim can be determined as true based solely on the given evidence.
    \item \textbf{REFUTES}: The claim can be determined as false based solely on the given evidence.
    \item \textbf{NOT ENOUGH INFORMATION (NEI)}: The claim cannot be judged as true or false using the provided evidence.
\end{itemize}

For the SUPPORTS and REFUTES labels, annotators write claims that rely only on the content of the evidence sentence, without using external knowledge or personal understanding. They are encouraged to reformulate, restructure, or vary the expression to enhance linguistic diversity and avoid simply paraphrasing the evidence. This approach differs from FEVER \cite{thorne-etal-2018-fever}, where claims often derive from minimal factual transformations. 

For the NEI label, annotators must still reuse at least one component from the evidence (subject, object, or event) while introducing new but contextually relevant information not present in the original sentence. This design ensures the NEI label remains semantically related to the evidence and also enables a more realistic evaluation of evidence retrieval and claim verification models.

Each evidence sentence leads to the creation of two distinct claim sentences for each label, resulting in a total of six claims per evidence. This procedure follows the approach in OCNLI \cite{ocnli} and IndoNLI \cite{mahendra-etal-2021-indonli}, where annotators are instructed to explore multiple information pieces within a single evidence and use diverse linguistic forms to represent similar meanings. Furthermore, for a given evidence sentence, the claims corresponding to different labels show clear variation in writing style, word choice, and focus, thereby improving the dataset’s semantic diversity.

During annotator training, we observed that a noticeable portion of annotators struggled with constructing SUPPORTS claims, especially when the evidence sentence contained limited information (typically 15–25 tokens). It was challenging for them to compose a natural claim without overly mirroring the evidence. Similarly, the REFUTES label posed difficulties due to excessive or unclear negation, which occasionally caused label drift toward NEI. 
\subsection{Corpus Validation}
\label{subsec:Corpus Validation}
To ensure data quality and annotation consistency, we perform a two-stage validation process: annotator validation and author validation, both conducted during the claim generation phase.

\subsubsection{Annotator Validation}
\label{subsubsec:Annotator Validation}
When 50\% of the corpus was completed, the top five annotators were selected to relabel a separate subset of 750 claim–evidence pairs that none had previously worked on. Labels were re-assigned collaboratively, and pairs with fewer than three matching labels were discarded. The inter-annotator agreement reached 99.87\% when at least three annotators agreed and 95.47\% when at least two agreed. The Fleiss $\kappa$ score was 95.87\%, significantly higher than FEVER’s 68.41\%, confirming strong consistency among annotators.

\subsubsection{Author Validation}
\label{subsubsec:Author Validation}
The same 750 pairs were further reviewed by the authors to assess guideline compliance. Results show that 96.12\% of the claims strictly followed the annotation rules, while only 3.87\% contained minor writing or labeling errors, demonstrating the overall reliability of the corpus.

\subsection{Corpus Analysis}
\label{subsec:Corpus Analysis}
\subsubsection{Overall Statistics}
\label{subsubsec:Overall Statistics}
We divide the corpus, which consists of 20,916 claim sentences, randomly into three different sets: 80.00\% for training (Train), 10.00\% for development (Dev), and 10.00\% for the test (Test) set for the VP task. For ER tasks, we do not employ pre-trained models on a portion of the data and then evaluate the remaining part. Instead, we use the cosine similarity measure between sentences within the model to extract evidence sentences, and this process will be carried out and evaluated on the entire corpus. The distribution of labels is summarized in Table \ref{tab:data-statistics}. 
\begin{table}[htb]
\centering
\begin{tabular}{lccc}
\hline
\multicolumn{1}{c}{\textbf{}} & \textbf{SUPPORTS} & \textbf{REFUTES} & \multicolumn{1}{l}{\textbf{NEI}} \\ \hline
\textbf{Train}      & 5,594 & 5,573 & 5,571 \\ \hdashline
\textbf{Dev}   & 666   & 694   & 730   \\ \hdashline
\textbf{Test}      & 708   & 706   & 677   \\ \hline
\end{tabular}
\caption{Overall statistics of our corpus.}
\label{tab:data-statistics}
\end{table}

\subsubsection{Length Distribution}
\begin{figure}
\begin{minipage}{0.45\textwidth}
    \includegraphics[width=1.0\linewidth]{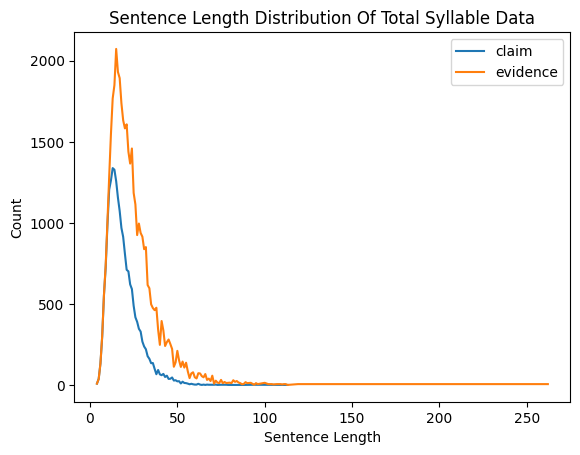}
    \caption{The distribution of total syllables in the corpus.}
    \label{fig:syllable_data_distribution}
\end{minipage}%
\hspace{0.05\textwidth}
\begin{minipage}{0.45\textwidth}
    \includegraphics[width=1.0\linewidth]{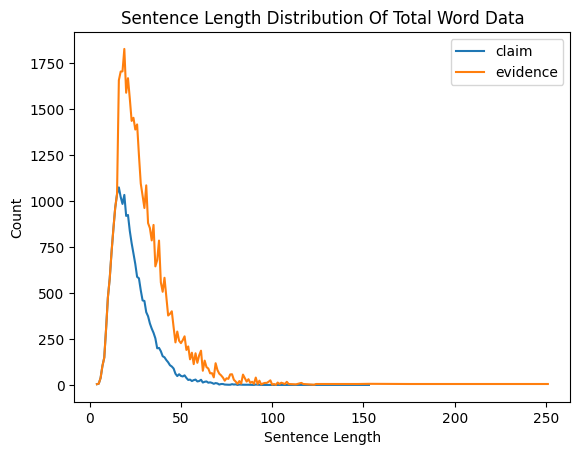}
    \caption{The distribution of total words in the corpus.}
    \label{fig:word_data_distribution}
\end{minipage}
\end{figure}

Allocate the sentences of claims and evidence based on their length, as presented in Figure \ref{fig:word_data_distribution} and Figure \ref{fig:syllable_data_distribution}. The length of claims is shorter compared to the evidence, but they still ensure a complete and accurate representation of the information conveyed by the evidence sentences. In reality, claim sentences often tend to be shorter in length, whereas evidence sentences are typically longer, serving the role of providing additional details and supporting the claim. Furthermore, writing too long sentences can result in providing more context and information to the model, making training and prediction less challenging and more straightforward. The shortest length for a claim is four words, while the longest when generating is 113 words. Sentences within the length range of 12 to 25 words constitute the largest proportion of the corpus. 

\subsubsection{New Word Rate}
\label{subsubsec:New Word rate}
\begin{table*}[h]
\centering
\resizebox{\textwidth}{!}{%
\begin{tabular}{lcccccccc}
\hline
\multicolumn{1}{c}{\multirow{2}{*}{\textbf{Label}}} & \multirow{2}{*}{\textbf{New Word Rate (\%)}} & \multirow{2}{*}{\textbf{New Dependency Rate (\%)}} & \multicolumn{6}{c}{\textbf{Part-Of-Speech (\%)}} \\ \cmidrule{4-9} 
 \multicolumn{3}{c}{}& \textbf{Noun} & {\textbf{Verb}} & {\textbf{Adjective}} & {\textbf{Preposition}} & {\textbf{Adjunct}} & {\textbf{Other}} \\ \hline
\textbf{SUPPORTS} & 32.91 & 75.26 & 26.97 & 30.22 & 7.65 & 10.81 & 7.92 & 16.45 \\ \hdashline
\textbf{REFUTES} & 31.28 & 67.90 & 26.27 & 25.47 & 8.66 & 8.04 & 13.10 & 18.47 \\ \hdashline
\textbf{NEI} & 50.44 & 81.96 & 33.96 & 23.61 & 8.96 & 9.82 & 7.65 & 15.99 \\ \hline
\end{tabular}%
}

\caption{Corpus analysis in terms of linguistic aspects.}
\label{tab:Corpus analysis in terms of linguistic aspects}
\end{table*}
To evaluate diversity in the corpus, we measure the rate of new words in the claim sentence that do not appear in the evidence sentence. We use VnCoreNLP \cite{vu-etal-2018-vncorenlp} for word segmentation. The results in table \ref{tab:Corpus analysis in terms of linguistic aspects} show that the new word rate in the REFUTES label is the lowest, reaching only 31.28\%. The word diversity of the NEI label (50.44\%) is the highest among the three labels, followed by the SUPPORTS label with 32.91\%. Both the SUPPORTS and REFUTES labels have a low new word rate compared to the NEI label, indicating that annotators tend to use fewer new words when constructing sentences for these two labels. This tendency ensures semantic accuracy with the evidence sentence when external knowledge is not used. 
In Addition, we utilized part-of-speech analysis on new words to delve deeper into annotator data construction trends. We used PhoNLP \cite{phonlp} for word segmentation and summarized the results in Table \ref{tab:Corpus analysis in terms of linguistic aspects}. The statistical results reveal that nouns and verbs are the two primary components annotators use when composing claim sentences.

\subsubsection{Corpus-generation rules analysis}
\label{sec:Corpus-generation rules analysis}
To evaluate and analyze the linguistic factors of the annotators during the creation of the corpus, we proceed with the analysis of the rules used based on the ViNLI \cite{huynh-etal-2022-vinli} corpus generation process. We randomly chose 500 SUPPORTS and 500 REFUTES claim-evidence pairs from the validation set. Selecting 500 sentences for each label will increase the coverage of the data compared to analyzing a smaller sample of only 100 sentences, as was performed in ViNLI. The statistical results and illustrative examples are provided in Appendix \ref{appendix:Data Generation Rules Analysis}.

For the SUPPORTS label, the rule annotator uses most is 'replace words with synonyms' with 58.60\%; meanwhile, the 'turn adjective into relative clause' rule has the lowest ratio (Table \ref{tab: support rules and example}). As for the REFUTES label, annotators tend to generate claims with the 'wrong reasoning about an event' rule, whereas 'opposite of time' and 'create a sentence that has the opposite meaning of a presupposition' share the same proportion of 5.60\%, making them the least common rule used when creating data (Table \ref{tab: refute rules and example}).  
\begin{figure}[htb]
\centering
\begin{minipage}{0.47\textwidth}
    \centering
    \includegraphics[height=0.5\linewidth]{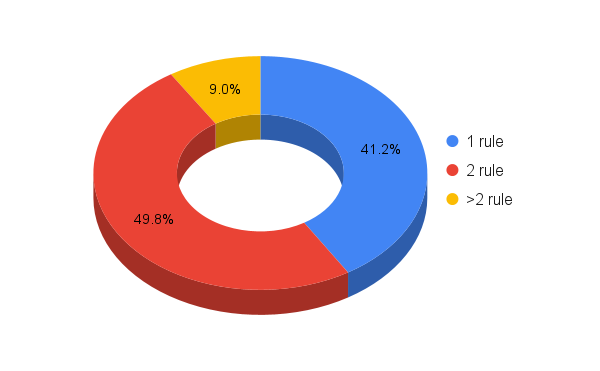}
    \caption{The ratio of SUPPORT claim generation rules.}
    \label{fig:SUPPORT claim rule6}
\end{minipage}%
\hspace{0.05\textwidth}
\begin{minipage}{0.47\textwidth}
    \centering
    \includegraphics[height=0.5\linewidth]{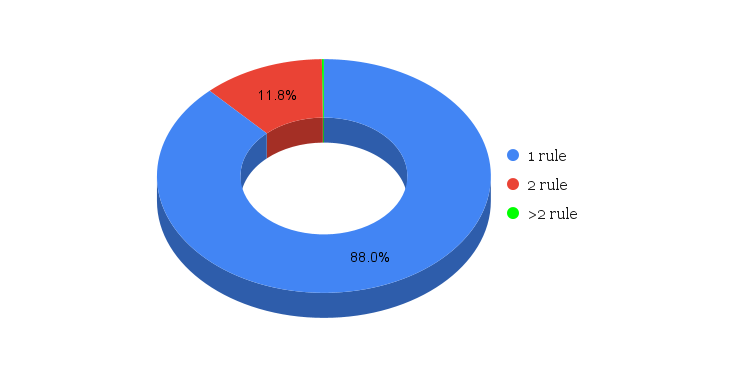}
    \caption{The ratio of REFUTE claim generation rules.}
    \label{fig:REFUTE claim rules}
\end{minipage}

\end{figure}

During the corpus generation process, annotators can use more than one rule for every claim they write. The analysis we conducted on claims shows that almost half of SUPPORT claims are written using two rules (49.80\%), while one rule has 41.20\%. For the REFUTES label, claims are nearly generated by one rule with a proportion of 88.00\% (Figure \ref{fig:REFUTE claim rules}). This can be explained by the fact that when using too many rules without reviewing carefully, REFUTES claims can be turned into NEI labels.

\section{Empirical Evaluation}
\label{sec:Empirical Evaluation}

\subsection{Baseline Models and Pipeline Description}
\label{subsec:Baseline Models and Pipeline Description}
To comprehensively evaluate the ViWikiFC dataset, we define two main subtasks: Evidence Retrieval (ER) and Verdict Prediction (VP), which are further integrated into a complete fact-checking pipeline.

\subsubsection{Evidence Retrieval}
\label{subsubsec:Evidence Retrieval}
In the ER task, given a claim sentence, the objective is to identify the most relevant evidence sentences from the corpus. We employ three retrieval methods: TF-IDF, Okapi BM25, and Vietnamese Sentence-BERT (SBERT). The TF-IDF model is implemented from scratch, BM25 is applied using the Rank-BM25\footnote{https://pypi.org/project/rank-bm25/} library with default hyperparameters, and Vietnamese Sentence-BERT is adopted to capture semantic similarity between claims and evidence through sentence embeddings. This model, based on the Sentence-BERT architecture, enables efficient comparison of sentence-level meanings with reduced computational cost compared to traditional transformer-based encoders.

\subsubsection{Verdict Prediction}
\label{subsubsec:Verdict Prediction}
For the VP task, we explore a range of models from traditional neural architectures to pre-trained language models. Simpler models, such as Continuous Bag of Words (CBOW) and BiLSTM, are trained with pre-trained PhoW2V embeddings at both syllable and word levels. We further investigate the performance of advanced transformer-based models, including Multilingual BERT (mBERT), XLM-RoBERTa, InfoXLM, PhoBERT, and ViDeBERTa, covering both multilingual and monolingual representations for Vietnamese. Word-based models (PhoBERT, ViDeBERTa) employ the VNCoreNLP toolkit for word segmentation, while syllable-based models (mBERT, XLM-R, InfoXLM) use raw tokenized text. To ensure robustness, each experiment is conducted five times with different random seeds, and the standard deviation of the performance metrics is reported.

\subsubsection{Pipeline Description}
\label{subsubsec:Pipeline Description}
Finally, we selected the models that achieved the highest performance in both tasks, evidence retrieval and verdict prediction, and integrated them into a complete fact-checking pipeline. This pipeline takes a claim sentence as input and outputs both an evidence sentence extracted from the corpus and a corresponding verdict label that indicates whether the claim is \textit{SUPPORTED}, \textit{REFUTED}, or \textit{NOT ENOUGH INFORMATION}. 

To ensure rigorous and reliable evaluation, we apply a \textbf{strict accuracy} metric, which considers a prediction correct only when both the retrieved evidence and the predicted label are accurate. In other words, if either the evidence or the verdict is incorrect, the entire prediction is counted as incorrect. This metric reflects the end-to-end quality of the fact-checking process rather than evaluating each component in isolation.

Strict accuracy is formally defined as follows, where $v$ and $v'$ denote the ground-truth and predicted evidence in the evidence retrieval task, and $e$ and $e'$ represent the ground-truth and predicted labels in the verdict prediction task \(e, e' \in \{SUPPORTS, REFUTES, NOT ENOUGH INFORM\\ATION\}\):

\begin{center}
Strict Accuracy = $ \delta(v, v') \times \delta(e, e')$
\end{center}

where $\delta(a, b)$ equals 1 if $a = b$ and 0 otherwise.

\subsubsection{Evaluation Metrics}
\label{subsubsec:Evaluation Metrix}
To assess the performance of the models in these tasks, we employ accuracy as the primary evaluation metric. Additionally, we utilize the F$_1$ score (macro average) as the secondary evaluation metric for the verdict prediction task (VP), which is described below:
\begin{center}
    F$_1$ = $\displaystyle \frac{2\ \times \ Precision\ \times \ Recall}{Precision\ +\ Recall}$
\end{center}

\subsection{Experimental Results}
\label{subsec:Experimental Results}
\begin{table*}[htb]
\centering
\fontsize{10pt}{13pt}\selectfont
\begin{tabular}{c||l|cccc}
\toprule \hline
\multicolumn{2}{c}{\multirow{2}{*}{\textbf{Model}}} & \multicolumn{2}{c}{\textbf{Dev}} & \multicolumn{2}{c}{\textbf{Test}} \\ \cmidrule{3-6}
                      \multicolumn{2}{c}{} &\textbf{Acc} & \textbf{F$_1$} & \textbf{Acc} & \textbf{F$_1$}   \\ \hline
\multirow{7}{*}{Word} & CBoW                 & 51.26$\pm{0.66}$      & 50.50$\pm{0.98}$     & 50.39$\pm{1.05}$      & 49.65$\pm{0.95}$       \\ \cdashline{2-6}
                      & BiLSTM               & 49.91 $\pm{0.47}$     & 49.83 $\pm{0.61}$     & 49.13 $\pm{0.45}$      & 49.24 $\pm{0.48}$      \\ \cdashline{2-6}
                      & ViDeBERTa$_{Small}$        & 61.67 $\pm{1.13}$     & 61.43 $\pm{0.79}$    & 60.62$\pm{0.66}$     & 61.73 $\pm{0.87}$      \\ \cdashline{2-6}
                      & ViDeBERTa$_{Base}$        & 64.41 $\pm{0.57}$     & 64.11 $\pm{0.54}$    & 63.70$\pm{0.51}$     & 64.06 $\pm{0.40}$      \\ \cdashline{2-6}
                      & PhoBERT$_{Base}$        & 81.72 $\pm{0.57}$     & 81.54 $\pm{0.58}$    & 80.67$\pm{0.37}$     & 80.70 $\pm{0.34}$      \\ \cdashline{2-6}
                      & PhoBERT$_{Base}$V2      & 82.50 $\pm{0.52}$       & 82.28 $\pm{0.53}$      & 81.22 $\pm{0.53}$      & 81.19 $\pm{0.56}$        \\ \cdashline{2-6}
                      & PhoBERT$_{Large}$      & \textbf{82.88 $\pm{0.97}$}       &\textbf{82.74 $\pm{1.01}$}      & \textbf{81.63 $\pm{0.47}$}      & \textbf{81.62 $\pm{0.44}$}        \\ \hline
\multirow{9}{*}{Syllable} &  CBoW &    45.25 $\pm{0.66}$ &  44.55 $\pm{0.63}$ &  44.24 $\pm{1.05}$ &  43.46 $\pm{0.64}$    \\ \cdashline{2-6}
                     & BiLSTM               & 47.68 $\pm{0.60}$     & 47.67 $\pm{0.77}$    & 45.12 $\pm{0.62}$     & 45.04 $\pm{0.79}$     \\ \cdashline{2-6}
                    & mBERT                & 75.75 $\pm{0.68}$     & 75.61 $\pm{1.24}$    & 75.93 $\pm{0.58}$     & 76.01 $\pm{1.18}$     \\ \cdashline{2-6}
                     & XLM-R$_{Base}$         & 79.66 $\pm{0.95}$     & 79.48 $\pm{0.97}$    & 78.53 $\pm{0.41}$     & 78.56 $\pm{0.48}$     \\ \cdashline{2-6}
                     & XLM-R$_{Large}$        & 86.09 $\pm{0.72}$     & 86.02 $\pm{0.67}$   & 85.11 $\pm{0.27}$    & 85.15 $\pm{0.26}$     \\ \cdashline{2-6}
                     & InfoXLM$_{Base}$         & 80.10 $\pm{0.57}$     & 79.92 $\pm{0.61}$    & 79.19 $\pm{0.78}$     & 79.23 $\pm{0.77}$     \\ \cdashline{2-6}
                     & InfoXLM$_{Large}$        & \textbf{87.45 $\pm{0.41}$}     & \textbf{87.42 $\pm{0.60}$}    & \textbf{86.50 $\pm{0.43}$}     & \textbf{86.51 $\pm{0.62}$}     \\ 
                     \hline \bottomrule
\end{tabular}
\caption{The result of models on Dev and Test.}
\label{tab:The result of models on Dev and Test}
\end{table*}

\subsubsection{Evidence Retrieval}
\label{subsubsec:Evidence Retrieval}
The experimental results presented in Table \ref{tab:Evidence Retrieval Results using accuracy} shed light on the performance of models in the evidence retrieval (ER) task with pre-trained model Vietnamese-SBERT (SBERT), TF-IDF, and BM25. These results offer valuable information on the strengths and weaknesses of these approaches.

We can observe that BM25 consistently outperforms TF-IDF and SBERT across all 3 labels in both single-sentence and 5-sentence retrieval scenarios. As the number of retrieved sentences increases from 1 to 5, the accuracy of all 3 models increases, with BM25 achieving the highest accuracy, reaching 93.93\% for the SUPPORTS label and 93.04\% for the REFUTES label. SBERT, with its transformer architecture, possesses the ability to capture the contextual nuances of the sentences it retrieves, but in the ViWikiFC corpus, the model's performance underperforms BM25. This suggests that BM25, despite its less complex architecture, exhibits remarkable performance in evidence retrieval tasks compared to other transformer-based methods \cite{beir}.

However, it is important to address the relatively lower performance of TF-IDF, SBERT, and BM25 in the NEI (Not Enough Information) label category. This discrepancy in performance can be attributed to the inherent challenges in dealing with the NEI label. On the NEI label, the information contained within a claim may be entirely unrelated to the available evidence. Therefore, accurately identifying and classifying such cases is inherently more complex. TF-IDF, SBERT, and BM25 struggle with this label due to the inherent ambiguity and lack of clear semantic connections between the claim and evidence.
\begin{table}[]
\centering
\resizebox{\linewidth}{!}{
\begin{tabular}{lcccccc}
\hline
\multicolumn{1}{c}{\multirow{2}{*}{\centering\textbf{ER}}} & \multicolumn{3}{c}{\textbf{Top 1}} & \multicolumn{3}{c}{\textbf{Top 5}} \\ \cmidrule{2-7} 
 & \textbf{SUP} & \textbf{REF} & \textbf{NEI} & \textbf{SUP} & \textbf{REF} & \textbf{NEI} \\  \midrule
TF-IDF & 61.47 & 58.22 & 34.26 & 83.67 & 81.46 & 54.58\\ \hdashline
SBERT & 80.19 & 71.57 & 43.61 & 89.41 & 84.69 & 60.10  \\ \hdashline 
BM25 & \textbf{88.30} & \textbf{86.93} & \textbf{56.57} & \textbf{93.93} & \textbf{93.04} & \textbf{71.51} \\ \hline
\end{tabular}
}   
\caption{Evidence Retrieval Results using accuracy (\%).}
\label{tab:Evidence Retrieval Results using accuracy}
\end{table}

\subsubsection{Verdict Prediction}
\label{sec:Verdict Prediction}
The analysis of the metrics clearly indicates that transformer-based models have emerged as the top performers in the task at hand. Their ability to capture complex linguistic patterns and relationships within the data sets them apart from other models.

In the case of the "word" corpus, PhoBERT$_{Large}$  stands out as the best-performing model. It achieves an impressive accuracy of 82.88\% on the development set and 81.63\% on the test set. These results demonstrate the robustness and effectiveness of PhoBERT$_{Large}$  in handling the nuances of the "word" corpus. Its high accuracy in both development and test sets suggests that it can consistently provide reliable predictions in this specific context.

However, when we shift our focus to the "syllable" corpus, InfoXLM$_{Large}$  takes the lead with remarkable accuracy and F$_1$ scores. It achieves an accuracy of 87.45\% and 86.50\% on validation and test sets, respectively. These numbers not only surpass the performance of PhoBERT$_{Large}$  but also highlight the exceptional capabilities of InfoXLM$_{Large}$ in handling the "syllable" corpus. This outcome establishes InfoXLM$_{Large}$ as the optimal model for the VP task. The superiority of InfoXLM$_{Large}$ in the "syllable" corpus might be attributed to its ability to capture the intricate linguistic patterns and syllable-level features, which are evidently critical in achieving high accuracy and F$_1$ scores in this particular context.

\subsubsection{Pipeline}
\label{sec:Pipeline}
We combine BM25 and SBERT, respectively, with the best models in the VP task, PhoBERT$_{Large}$  and InfoXLM$_{Large}$  for VP. The pipeline is evaluated through the test set. 
\begin{table}[htb]
\centering
\resizebox{\linewidth}{!}{
\begin{tabular}{c||cccc}
\hline
\multicolumn{2}{c}{\textbf{Pipeline}} & \textbf{ER Acc} & \textbf{VP Acc} & \textbf{Strict Acc} \\ \hline
\multirow{2}{*}{SBERT} &  InfoXLM$_{Large}$   & 65.56   & 75.80  & 57.15 \\ \cdashline{2-5}
\multirow{1}{*}{}&PhoBERT$_{Large}$ & 65.56   & 71.54  & 52.84           \\ \hline
\multirow{2}{*}{BM25} & InfoXLM$_{Large}$   & 78.38   & 80.63  & 67.00 \\ \cdashline{2-5}
\multirow{1}{*}{}&PhoBERT$_{Large}$  & 78.38   & 76.18  & 63.46           \\ \hline
\end{tabular}
}
\caption{Pipeline result for test set}
\label{tab:Pipeline result for test set}
\end{table}
Table \ref{tab:Pipeline result for test set} gives the result of the pipeline; the best combination for both tasks is BM25 and InfoXLM$_{Large}$  with 67.00\%, while SBERT and PhoBERT$_{Large}$  have the lowest result. This result demonstrates the ViWikiFC corpus's challenge for current Vietnamese fact-checking models. Besides, it provides a practical perspective on the fact-checking problem, as accurate evidence retrieval significantly influences the performance of evidence-based methods such as PhoBERT and InfoXLM. While these methods perform well in verdict prediction tasks, their strict accuracy still needs to be improved. Therefore, approaching the Vietnamese fact-checking problem using separate models for each task, first for evidence retrieval and then for verdict prediction, needs improvement to deal with more complex claims that have been studied in English \cite{pan-etal-2023-fact,yao2023end}.

\section{Conclusions and Future Directions}
\label{sec:Conclusions and Future Directions}
In this paper, we construct the first open-domain and high-quality Wikipedia Vietnamese corpus, including 20,916 samples to evaluate fact-checking models. Furthermore, we have discussed the data collection and annotation method and shared the insight we obtained during the annotation process, which can be applied in the development of non-English corpus creation. The best baseline and pipeline model (InfoXLM and BM25) only achieved 67.00\% strict accuracy, making it a feasible challenge for the Vietnamese language model in fact-checking tasks. We believe that ViWikiFC will encourage the development of Vietnamese fact-checking research.

Following the development of fact-checking in English, we aim to develop our research for Vietnamese fact-checking by expanding our corpus from quantity to quality with more trustworthy data sources and constructing datasets not only on textual but also on images and tables for developing multi-modal fact-checking. Besides, we want our work to adapt to more advanced NLP tasks in low-resource languages, such as fake news detection or machine reading comprehension, with advanced reasoning design and sub-task functionalities.

\section*{Acknowledgement}
This research is funded by Vietnam National University Ho Chi Minh City (VNU-HCM) under the grant number DS2025-26-01.

\section*{Declarations}

\textbf{Conflict of interest} The authors declare that they have no conflict of interest.

\section*{Data Availability}

Data available on reasonable request.

\section*{Author Contribution}
Hung Tuan Le: Conceptualization; Formal analysis; Investigation; Methodology; Validation; Visualization; Writing - review\&editing.
Long Truong To: Conceptualization; Data curation; Formal analysis; Investigation; Validation; Visualization; Writing - original draft.
Manh Trong Nguyen: Conceptualization; Data curation; Investigation; Methodology; Writing - original draft.
Kiet Van Nguyen: Conceptualization; Formal analysis; Investigation; Methodology; Validation; Supervision; Writing - review\&editing.

\section{Bibliographical References}
\bibliographystyle{lrec2026-natbib}
\bibliography{lrec2026-example}
\newpage
\appendix
\onecolumn
\section{Data Generation Rules Analysis}
\label{appendix:Data Generation Rules Analysis}

\begin{longtable}{|p{2cm}|p{9cm}|l|}

\hline
\multicolumn{1}{|c|}{Rule} &
  \multicolumn{1}{c|}{Example} &
  \multicolumn{1}{c|}{Ratio} \\ \hline
\endfirsthead

\hline
\multicolumn{1}{|c|}{Rule} &
  \multicolumn{1}{c|}{Example} &
  \multicolumn{1}{c|}{Ratio} \\ \hline
\endhead
\label{tab: support rules and example}
Change active sentences into passive sentences and vice versa &
  \textbf{E}: Âu Lạc {\textcolor{blue}{bị}} nhà Triệu ở phương Bắc {\textcolor{blue}{thôn tính}} vào đầu thế kỷ thứ 2 TCN sau đó là thời kỳ Bắc thuộc kéo dài hơn một thiên niên kỷ. (\textit{Au Lac {\textcolor{blue}{was conquered}} by the Zhaos in the northern region at the beginning of the 2nd century BC, leading to the Northern Domination period that lasted for over a millennium.})
  
  \textbf{C}: Thời kỳ Bắc thuộc diễn ra sau khi phương Bắc {\textcolor{blue}{thôn tính}} được Âu Lạc. (\textit{The Northern Domination period took place after the northern regions {\textcolor{blue}{conquered}} Au Lac.})&
  \multirow{5}{*}{17.40\%} \\ \hline
  Replace words with synonyms. &
  \textbf{E}: Thời kỳ đầu, {\textcolor{red}{những bậc đế vương và những nhà quý tộc}} của La Mã {\textcolor{blue}{thích}} lụa Trung Hoa đến mức họ cho cân lụa lên và đổi chỗ lụa đó bằng vàng với cân nặng tương đương. (\textit{In the early periods, Roman {\textcolor{red}{emperors and nobles}} {\textcolor{blue}{had such a fondness}} for Chinese silk that they would weigh silk against gold and exchange it with an equivalent weight of gold.})

  \textbf{C}: Lụa Trung Hoa {\textcolor{blue}{nhận được sự ưa chuộng cực lớn}} từ {\textcolor{red}{những người đứng đầu, nhũng người có địa vị}} ở La Mã. (\textit{Chinese silk {\textcolor{blue}{received immense favor}} from the Roman {\textcolor{red}{leaders and influential people}}.})
  &
  \multirow{5}{*}{58.60\%} \\ \hline
  Add or remove modifiers that do not radically alter the meaning of the sentence. &
  \textbf{E}: {\textcolor{blue}{Gary Shilling, chủ tịch một công ty nghiên cứu kinh tế}}, cho rằng mức tăng trưởng GDP thực sự của Trung Quốc chỉ là 3.50\% chứ không phải 7.00\% như báo cáo chính thức. (\textit{{\textcolor{blue}{Gary Shilling, the chairman of an economic research firm}}, believes that China's actual GDP growth rate is only 3.50\%, not the officially reported 7.00\%.})
  
  \textbf{C}: Báo cáo chính thức về mức tăng trưởng thực sự của Trung Quốc đã được {\textcolor{blue}{Gary Shilling}} cho rằng là 3.50\%. (\textit{The official report on China's actual growth rate has been suggested by Gary Shilling to be 3.50\%.})&   
  \multirow{5}{*}{48.00\%} \\ \hline
  Turn nouns into relative clauses &
  \textbf{E}: Trong bốn nước xưa có nền văn minh lớn thì có ba nước xưa ở vào {\textcolor{blue}{châu Á}} (Ấn Độ, Iraq (Lưỡng Hà) và Trung Quốc). (\textit{Among the four ancient civilizations, three were located in {\textcolor{blue}{Asia}} (India, Iraq (Mesopotamia), and China).})

  \textbf{C}: {\textcolor{blue}{Châu Á là nơi}} tồn tại ba nước có nền văn minh lớn. (\textit{{\textcolor{blue}{Asia is the region where}} three major ancient civilizations existed.}) 
  &
  \multirow{5}{*}{10.20\%} \\ \hline
  Turn the object into relative clauses &
  \textbf{E}: Lý thuyết "Quả cầu tuyết Trái Đất" cho rằng những sự thay đổi về {\textcolor{blue}{mức độ CO2}} vừa là nguyên nhân gây ra, vừa là nguyên nhân làm kết thúc thời kỳ cực lạnh ở cuối Liên đại Nguyên Sinh (Proterozoic). (\textit{The "Snowball Earth" theory posits that changes in {\textcolor{blue}{the levels of CO2}} were both the cause and the termination of the extreme cold period at the end of the Proterozoic Eon.})
  
  \textbf{C}: Theo lý thuyết "Quả cầu tuyết Trái Đất" cho thấy rằng sự thay đổi thành phần không khí mà cụ thể là {\textcolor{blue}{lưu lượng CO2 là tác nhân}} cho kỷ băng giá, và cũng kết thúc luôn thời kỳ cực lạnh ở kỷ Proterozoic. (\textit{The "Snowball Earth" theory indicates that changes in the composition of the atmosphere, specifically {\textcolor{blue}{the levels of CO2 which are the driving factors}} behind ice ages and the termination of extreme cold periods in the Proterozoic Eon.})
  &
 \multirow{10}{*}{2.20\%} \\ \hline
 Turn adjectives into relative clauses  &
 \textbf{E}: Về du lịch biển, Nghệ An có 82 km bờ biển với {\textcolor{blue}{nhiều bãi tắm đẹp hấp dẫn}} khách du lịch quốc tế như bãi biển Cửa Lò, Cửa Hội; Nghi Thiết,... (\textit{For beach tourism, Nghệ An has a coastline of 82 kilometers with {\textcolor{blue}{numerous beautiful and attractive beaches}} for international tourists, such as Cua Lo, Cua Hoi, Nghi Thiet,...})
 
 \textbf{C}: Nghệ An có {\textcolor{blue}{các bãi tắm, nơi được xem là những địa điểm  tuyệt đẹp và rất cuốn hút}} du khách quốc tế. (\textit{Nghe An has {\textcolor{blue}{beaches which are considered to be stunning and highly attractive destinations}} for international tourists.}) &
 \multirow{8}{*}{0.20\%}  \\ \hline
 Replace quantifiers or time with others that have a similar meaning.
 &
 \textbf{E}: Theo Viện Hàn lâm Khoa học Nga, Liên Xô đã chịu {\textcolor{blue}{26,6 triệu}} thương vong trong chiến tranh thế giới thứ hai, bao gồm sự gia tăng tỷ lệ tử vong ở trẻ sơ sinh là 1,3 triệu. (\textit{According to the Russian Academy of Sciences, the Soviet Union suffered {\textcolor{blue}{26.6 million}} casualties during World War II, including an increase in the infant mortality rate of 1.3 million.}) 

 \textbf{C}: Liên Xô chứng kiến {\textcolor{blue}{gần 27 triệu}} cái chết trong thế chiến thứ hai. (\textit{The Soviet Union witnessed {\textcolor{blue}{nearly 27 million}} deaths during World War II.})
   & \multirow{8}{*}{5.60\%} \\ \hline
   Create a presupposition sentence &
   \textbf{E}: Một số người cho rằng Firenze trở thành nơi khởi đầu Phục Hưng là do may mắn, nghĩa là đơn thuần bởi vì những vĩ nhân ngẫu nhiên sinh ra ở đây: cả da Vinci, Botticelli và {\textcolor{blue}{Michelangelo đều là người xứ Toscana (mà Firenze là thủ phủ)}}. (\textit{Some argue that Firenze became the cradle of the Renaissance due to luck, meaning it was purely coincidental that such great talents were born there: both da Vinci, Botticelli, and {\textcolor{blue}{Michelangelo were natives of Tuscany (with Firenze as its capital)}}.})   
   
   \textbf{C}: {\textcolor{blue}{Quê hương của Michelangelo là Firenze.}} (\textit{{\textcolor{blue}{Michelangelo's hometown is in Firenze.}}})
   &
   \multirow{8}{*}{20.60\%}   \\ \hline
   Replace Named Entities with a word that stands for the class. &
   \textbf{E}: {\textcolor{blue}{Các công ty công nghệ cao của Trung Quốc như Lenovo, Huawei, Xiaomi, Coolpad, ZTE,...}} đã bắt đầu cạnh tranh thành công trên thị trường thế giới. (\textit{{\textcolor{blue}{High-tech companies in China such as Lenovo, Huawei, Xiaomi, Coolpad, ZTE,...}} have successfully begun competing in the global market.})

   \textbf{C}:  {\textcolor{blue}{Nhiều công ty công nghệ của Trung Quốc}} đã bắt đầu ghi danh mình thành công trên thị trường công nghệ cao thế giới. (\textit{{\textcolor{blue}{Many Chinese high-tech companies}} have indeed successfully made a name for themselves in the global high-tech market.}) &
   \multirow{8}{*}{2.00\%}         \\ \hline 
      Create conditional sentences &
   \textbf{E}: Bệnn viêm gan siêu vi C mạn được xác định là nhiễm siêu vi viêm gan C hơn 6 tháng căn cứ trên sự hiện diện của ARN. (\textbf{The diagnosis of chronic hepatitis C infection is established when the presence of RNA is detected for more than 6 months.})

   \textbf{C}:  {\textcolor{blue}{Nếu}} nhiễm siêu vi gan C hơn nửa năm, bạn {\textcolor{blue}{sẽ}} bị viêm gan siêu vi C mạn. (\textit{{\textcolor{blue}{If}} you are infected with the hepatitis C virus for more than half a year, you will be considered to have chronic hepatitis C.}) &
   \multirow{8}{*}{0.40\%}         \\ \hline 
   Other  &
   \textbf{E}: Những phân tích lõi băng và lõi trầm tích đại dương không chứng minh rõ ràng sự hiện diện của băng giá và những thời kỳ trung gian băng giá trong vòng vài triệu năm qua. (\textit{Core ice and sediment analyses from the ocean do not provide clear evidence of the presence of ice ages and intermediate ice-free periods over the past several million years.})

   \textbf{C}:  {\textcolor{blue}{Dù}} đều là các bằng chứng địa chất nhưng theo những phân tích lõi băng và lõi trầm tích đại dương đã không chứng minh rõ ràng sự hiện diện của băng giá và những thời kỳ trung gian băng giá trong vòng vài triệu năm qua. (\textit{{\textcolor{blue}{Although}} both are geological pieces of evidence, core ice and ocean sediment analyses have not provided clear confirmation of the presence of ice ages and intermediate ice-free periods over the past several million years.})
   
   \textbf{Note}: Although clause, Inversion can be used for converting evidence to claim. These circumstances are listed in Other.
   &
   \multirow{15}{*}{3.00\%}         \\ \hline
    \caption{SUPPORTS rules and examples for creating evidence (E) - claim (C) pairs. Simply, we only mention one rule to be applied in each example.}

\end{longtable}

\begin{longtable}{|p{2cm}|p{9cm}|l|}

\hline
\multicolumn{1}{|c|}{Rule} &
  \multicolumn{1}{c|}{Example} &
  \multicolumn{1}{c|}{Ratio} \\ \hline
\endfirsthead

\hline
\multicolumn{1}{|c|}{Rule} &
  \multicolumn{1}{c|}{Example} &
  \multicolumn{1}{c|}{Ratio} \\ \hline
\endhead
\label{tab: refute rules and example}
Use negative words (no, not, never, nothing, hardly, etc.) &
  \textbf{E}: Sự kiện này dẫn tới việc Hiệp định Genève (1954) được ký kết và Việt Nam bị chia cắt thành hai vùng tập kết quân sự, lấy ranh giới là vĩ tuyến 17. (\textbf{This event led to the signing of the Geneva Accords in 1954, which divided Vietnam into two military zones along the 17th parallel.})
  
  \textbf{C}: Sự kiện này dẫn tới việc Hiệp định Genève (1954) {\textcolor{red}{không}} được ký kết. (\textit{This event led to the Geneva Accords (1954) {\textcolor{red}{not}} being signed.}) &
  \multirow{5}{*}{23.00\%} \\ \hline
  Replace words with antonyms &
  \textbf{E}:Đội tuyển bóng nước Singapore đã giành huy chương vàng SEA Games lần thứ 27 vào năm 2017, tiếp tục chuỗi {\textcolor{blue}{vô địch}} dài nhất của thể thao Singapore về môn môn cụ thể. (\textit{The Singapore water polo team won the gold medal at the 27th SEA Games in 2017, continuing Singapore's longest {\textcolor{blue}{winning}} streak in a specific sport.})

  \textbf{C}:Singapore liên tục gặp {\textcolor{red}{thất bại}} ở bộ môn bóng nước tại các kỳ SEA Games. (\textit{Singapore has continuously faced {\textcolor{red}{defeats}} in the sport of water polo at various SEA Games.})
  &
  \multirow{5}{*}{14.00\%} \\ \hline
  Opposite of quantity &
  \textbf{E}: Năm 2010, tổng chi tiêu của Nhà nước vào khoa học và công nghệ chiếm khoảng {\textcolor{blue}{0.45\%}} GDP. (\textit{In 2010, the government's total expenditure on science and technology accounted for approximately {\textcolor{blue}{0.45\%}} of the GDP.})
  
  \textbf{C}: Nhà nước chi ra hơn {\textcolor{red}{2.00\%}} GDP năm 2010 cho khoa học và công nghệ. (\textit{The government allocated over {\textcolor{red}{2.00\%}} of the GDP in 2010 for science and technology.})&   
  \multirow{5}{*}{9.80\%} \\ \hline
  Opposite of time &
  \textbf{E}: {\textcolor{blue}{Năm 1860}}, dân số Singapore đã vượt quá 80,000 và hơn một nửa là người Hoa. (\textit{In {\textcolor{blue}{1860}}, the population of Singapore had exceeded 80,000, with over half being of Chinese descent.})

  \textbf{C}: Mãi đến {\textcolor{red}{năm 2000}}, Dân số Singapore mới đạt mốc 80,000. (\textit{It wasn't until {\textcolor{red}{the year 2000}} that Singapore's population reached the 80,000 milestone.}) 
  &
  \multirow{5}{*}{5.60\%} \\ \hline
  Create a sentence that has the opposite meaning of a presupposition &
  \textbf{E}: Đức đã tung ra {\textcolor{blue}{70.00\% binh lực với các sư đoàn mạnh và tinh nhuệ nhất, chưa kể binh lực góp thêm của các nước đồng minh của Đức (Ý, Rumani, Bulgari, Hungary, Phần Lan...)}}. (\textit{Germany deployed {\textcolor{blue}{approximately 70.00\% of its military power, not counting the additional military contributions from Germany's allies (Italy, Romania, Bulgaria, Hungary, Finland...)}}.})
  
  \textbf{C}: Đức {\textcolor{red}{không có bất kỳ đồng minh nào giúp sức}}. (Germany {\textit{\textcolor{red}{did not have any allies providing assistance}}.})
  &
 \multirow{8}{*}{5.60\%} \\ \hline
 Wrong reasoning about an object (House, car, river, sea, person, etc.)  &
 \textbf{E}: Khi gió mùa đổi hướng, các đường bờ biển giáp với {\textcolor{blue}{biển Ả Rập và vịnh Bengal}} có thể phải hứng chịu xoáy thuận. (\textit{When the monsoon winds change direction, coastal areas bordering {\textcolor{blue}{the Arabian Sea and the Bay of Bengal}} may be susceptible to cyclones.})
 
 \textbf{C}: Các đường bờ biển tiếp giáp {\textcolor{red}{Ấn Độ}} có thể phải hứng chịu xoáy thuận khi gió mùa đổi hướng. (\textit{The coastal areas bordering {\textcolor{red}{India}} may be susceptible to cyclones when the monsoon winds change direction.}) &
 \multirow{8}{*}{23.60\%}  \\ \hline
 Wrong reasoning about an event &
 \textbf{E}: Một trận động đất ở Valdivia, Chile với cường độ 9,4-9,6 độ richter, mức cao nhất từng được ghi nhận, {\textcolor{blue}{khiến 1.000 đến 6.000 người chết}}. (\textit{An earthquake in Valdivia, Chile, with a magnitude of 9.4 to 9.6 on the Richter scale, the highest ever recorded, {\textcolor{blue}{resulted in the deaths of approximately 1,000 to 6,000 people}}.}) 

 \textbf{C}: Cơn địa chấn ở Valvidia chỉ {\textcolor{red}{gây ra thiệt hại về tài sản}}. (\textit{The earthquake in Valdivia only {\textcolor{red}{caused property damage}}.})
   & \multirow{8}{*}{30.80\%} \\ \hline
   Others &
   \textbf{E}: Tiếng Pháp không phải là ngôn ngữ chính thức ở Ontario, nhưng Đạo luật Dịch vụ Ngôn ngữ Pháp đảm bảo rằng các dịch vụ của tỉnh bang sẽ được cung cấp bằng ngôn ngữ này. (\textit{French is not the official language in Ontario, but the French Language Services Act ensures that provincial services will be provided in this language.})   
   
   \textbf{C}: {\textcolor{red}{Dù}} tiếng Pháp có là ngôn ngữ chung của bang Ontario,  {\textcolor{red}{thì}} nó cũng sẽ bị loại bỏ khỏi hệ thống các dịch vụ. (\textit{{\textcolor{red}{Even though}} French is one of the official languages of the province of Ontario, it will be removed from the system of services.})
   &
   \multirow{8}{*}{0.20\%}   \\ \hline

    \caption{REFUTES rules and examples for creating evidence (E) - claim (C) pairs. , we only mention one rule to apply in each example.}

\end{longtable}

\end{document}